\documentclass[letterpaper, 10 pt, conference]{ieeeconf}  

\IEEEoverridecommandlockouts                              

\usepackage[noadjust]{cite}

\usepackage{graphicx} 
\usepackage{subfigure}
\usepackage{amsmath} 
\usepackage{amssymb}  
\usepackage{multirow}
\usepackage{upgreek}
\usepackage{mathtools,lipsum, nccmath}
\usepackage{ulem}
\usepackage{xcolor}

\title{\LARGE \bf
A Morphing Quadrotor that Can Optimize Morphology for Transportation}

\author{Chanyoung Kim$^{1}$, Hyungyu Lee$^{2}$, Myeongwoo Jeong$^{1}$, and Hyun Myung$^{2}$, \textit{Senior Member, IEEE}
\thanks{$^{1}$Chanyoung Kim and Myeongwoo Jeong are with School of Electrical Engineering, Korea Advanced Institute of Science and Technology (KAIST), Daejeon, 34141, Republic of Korea.}%
\thanks{$^{2}$Hyungyu Lee and Hyun Myung are with Urban Robotics Lab, School of Electrical Engineering, KI-AI, KI-R, KAIST, Daejeon, 34141, Republic of Korea.}
\thanks{This research was supported by the National Research Foundation of Korea (NRF) Grant funded by the Ministry of Science and ICT for First-Mover Program for Accelerating Disruptive Technology Development (NRF-2018M3C1B9088328). The students are supported by the BK21 FOUR from the Ministry of Education (Republic of Korea).}
}%

\begin{document}

\maketitle
\thispagestyle{empty}
\pagestyle{empty}

\begin{abstract}
Multirotors can be effectively applied to various tasks, such as transportation, investigation, exploration, and lifesaving, depending on the type of payload. However, due to the nature of multirotors, the payload loaded on the multirotor is limited in its position and weight, which presents a major disadvantage when the multirotor is used in various fields. In this paper, we propose a novel method that greatly improves the restrictions on payload position and weight using a morphing quadrotor system. Our method can estimate the drone's weight, center of gravity position, and inertia tensor in real-time, which change depending on payload, and determine the optimal morphology for efficient and stable flight. An adaptive control method that can reflect the change in flight dynamics by payload and morphing is also presented. Experiments were conducted to confirm that the proposed morphing quadrotor improves the stability and efficiency in various situations of transporting payloads compared with the conventional quadrotor systems.
\end{abstract}

\section{Introduction} \label{sec1}

Multirotors can be effectively applied to various tasks, depending on the type of payload, based on their superior mobility and hovering ability to move freely in 3D space. If parcels are loaded, a multirotor can be used as a delivery drone. If a camera or LiDAR is loaded, it can be used as an investigation and exploration drone \cite{Ashour}, \cite{ Jung}, and if it loads rescue equipment, it can be used as a lifesaving drone \cite{Xiang}. Furthermore, multirotors can perform a number of complex tasks using robotic arms \cite{Ohnishi} or grasping units \cite{Lindsey}.

However, due to the nature of multirotors, the efficiency and stability change depending on the payload's position and weight, and sometimes this change causes the drone to fall. Therefore, the multirotor's payload is limited in its position and weight, which presents a significant disadvantage when the multirotor is used in various fields. For example, when carrying a relatively heavy payload such as a delivery package or construction material, the payload should be placed at the center of the drone. Even if the payload is loaded at the center, the center of gravity of the payload cannot always be at the center, and thus, efficiency and stability degradation can occur. Moving equipment, such as a robotic arm, or equipment that should be mounted far from the center of the drone, such as a contact-based inspection sensor \cite{Ollero}, \cite{Ikeda}, must be light to minimize the change in the drone's center of gravity. Increasing the size of the drone can be considered as a solution to reduce the restrictions on the payload. However, because increasing the size of the drone increases the cost and damage in case of accidents and reduces the accessible area, it is not an optimal solution.

    \begin{figure}[t] 
    \centering
        \includegraphics[width=8.5cm]{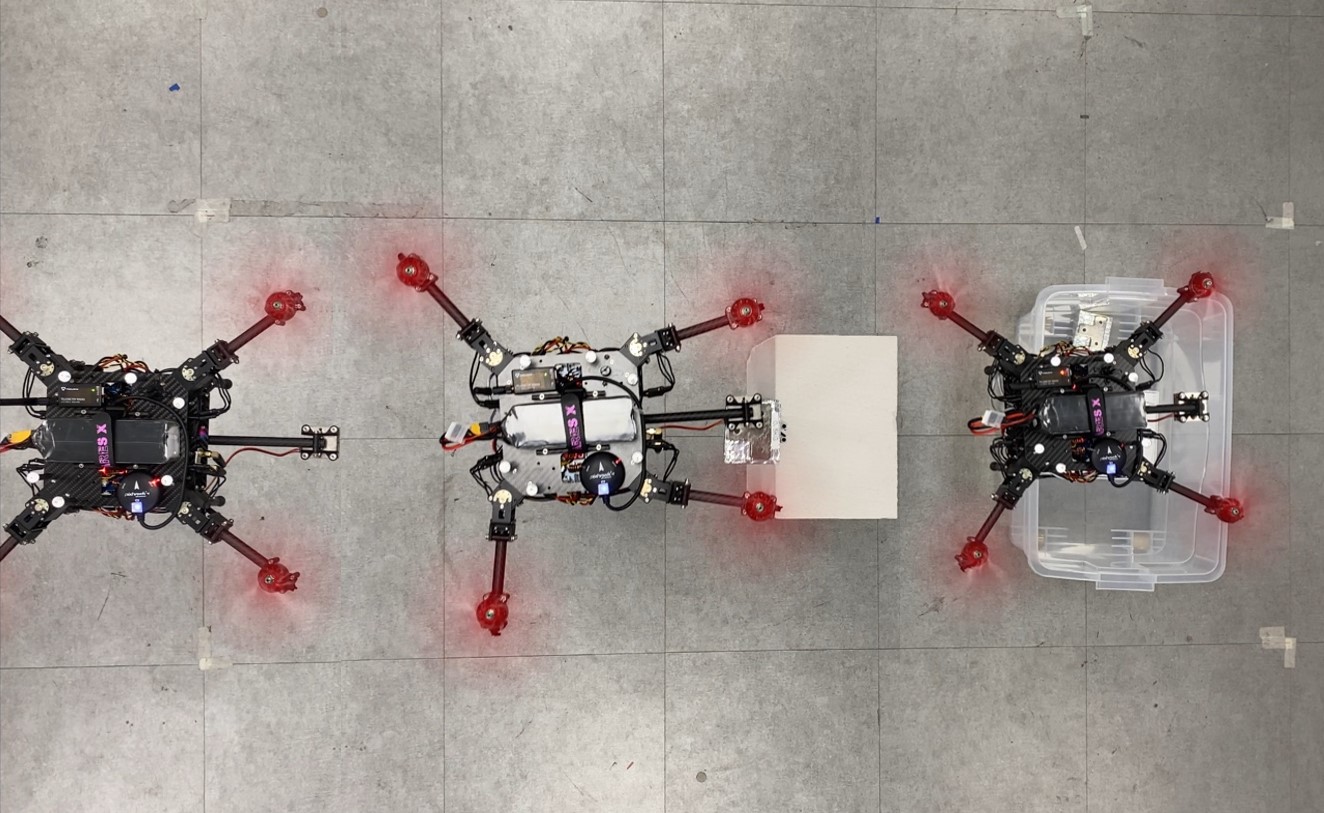}
        \caption{Our morphing quadrotor performing an aerial grasping and dropping task.}
        \label{fig_title}
    \end{figure}

Meanwhile, to overcome the hardware limitation of multirotors and increase the utility of multirotors, many studies on tilting or morphing multirotors have been conducted. In Voliro \cite{Kamel} and TiltDrone \cite{Zheng}, 6-DOF control was performed by tilting rotors, whereas only 4-DOF control is possible in general multirotors due to coupling of rotational motion and translational motion. The wall climbing quadrotor that can land vertically on a wall by tilting its rotor was studied in \cite{WC}, \cite{WC2}. Morphing multirotors, which fold the frame when passing through narrowly spaced obstacles, have also been studied \cite{Zhao, Riviere, Bucki, Yang, Falanga}. In particular, a foldable drone with four arms that can independently rotate around the main body was presented by Falanga \textit{et al}. \cite{Falanga}. Improving energy efficiency with a morphing drone that can rotate its four arms around the center of the drone was studied by Xiong \textit{et al}. \cite{Xiong}. However, in \cite{Xiong}, the optimal morphology for transporting the payload was not uniquely determined. In addition, to increase energy efficiency, information on the weight and the center of gravity of the drone, which change depending on payload, was required. Also, flight dynamics changed by payload were not reflected in controlling the drone.

In this paper, we propose a novel method that can greatly improve the limit on the position and weight of payload by using the quadrotor with the morphing form proposed in \cite{Falanga}. Our method can estimate the drone's weight, center of gravity position, and inertia tensor in real-time, which change depending on payload, and determine the optimal morphology uniquely for efficient and stable flight. An adaptive control method that can reflect the change in flight dynamics by payload and morphing is also presented. Experiments were conducted to confirm that the proposed morphing quadrotor improves the stability and efficiency in various situations of transporting payloads compared with the conventional quadrotor systems (Refer to Fig. \ref{fig_title}).

\section{Modeling} \label{sec2}

    \begin{figure}[t] 
    \centering
        \includegraphics[scale=0.29]{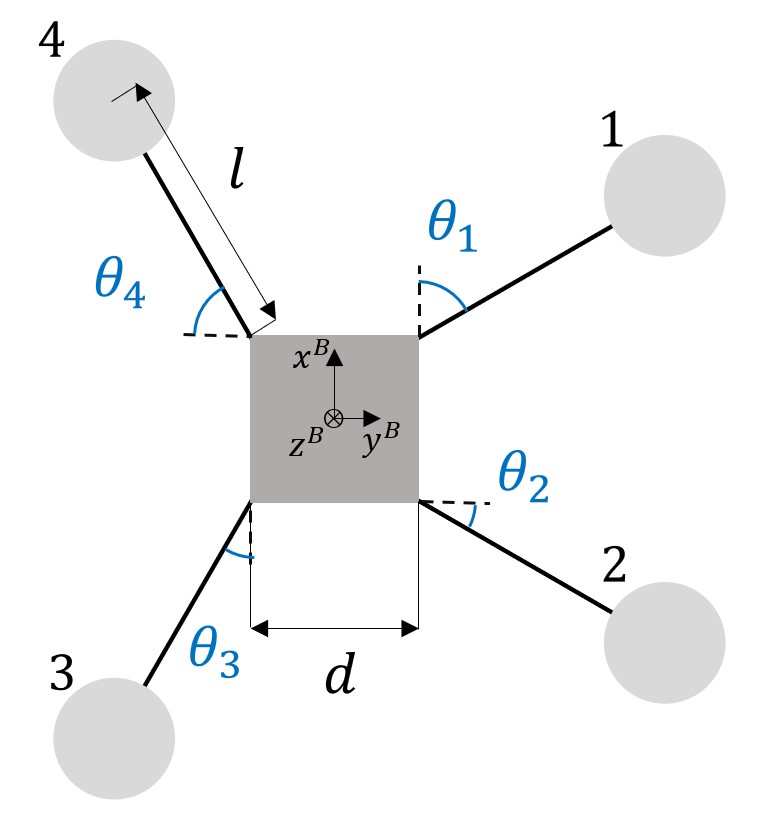}
        \caption{Schematics of the morphing quadrotor}
        \label{fig_schematics}
    \end{figure}

As shown in Fig. \ref{fig_schematics}, the morphing quadrotor consists of a main body of width $d$ and four arms of length $l$. The four arms can independently adjust the angle $\theta_i$ $( i=1,2,3,4)$ around the $z$-axis of the body frame where the $z$-axis is set downward \cite{Falanga}. The $x$-coordinate $x_i$ $(i=1,2,3,4)$ and $y$-coordinate $y_i$ $(i=1,2,3,4)$ of the rotor of the morphing quadrotor are calculated as follows:

    \begin{equation}
    \label{eq1}
    \begin{aligned}
        (x_1, y_1) & = (  d / 2 + l \cos \theta_1,   d / 2 + l \sin \theta_1),
        \\
        (x_2, y_2) & = (- d / 2 - l \sin \theta_2,   d / 2 + l \cos\theta_2),
        \\
        (x_3, y_3) & = (- d / 2 - l \cos \theta_2, - d / 2 - l \sin \theta_3),
        \\
        (x_4, y_4) & = (  d / 2 + l \sin \theta_4, - d / 2 - l \cos \theta_4).
        \\
    \end{aligned}
    \end{equation}

The thrust and torque generated by the rotor are modeled to be proportional to the square of the rotor's rotational speed as follows:

    \begin{equation}
    \label{eq2}
    \begin{aligned}
        f_i 
        =
        \mu {\Omega_i}^2,
        \\
        \tau_i
        =
        \kappa {\Omega_i}^2,
    \end{aligned}
    \end{equation}

\noindent
where $f_i$ and $\tau_i$ are the thrust force and torque generated by the $i$-th rotor, respectively. $\mu$ is the lift force coefficient, and $\kappa$ is the drag torque
coefficient.

The relationship between the thrust of the rotor and the power consumption is modeled as follows, according to \cite{Leishman}:

    \begin{equation}
    \label{eq3}
        P_i 
        =
        \gamma \sqrt{\frac{f_{i}^{3}}{2 \rho S}} ,
    \end{equation}

\noindent
where $P_i$ stands for the power consumption of the $i$-th rotor, $\gamma$ is the loss factor determined experimentally, $S$ is the disk area of the propeller, and $\rho$ is the air density.

The forces and moments that the thrust and torque of the four rotors act on the drone are as follows:
    
    \begin{equation}
    \label{eq4}
        \mathbf{F}^{B}
        =
        \sum_{i=1}^{4} - f_{i} \mathbf{e}_{z}^{B},
    \end{equation}
    
    \begin{equation}
    \label{eq5}
    \begin{aligned}
        \mathbf{M}^{B}
        =
        \sum_{i=1}^{4} - \left((x_{i}-x_{\mathrm{CoG}}) \mathbf{e}_{x}^{B} + (y_{i}-y_{\mathrm{CoG}}) \mathbf{e}_{y}^{B}\right) &\\
        \times f_{i} \mathbf{e}_{z}^{B} + \sum_{i=1}^{4} \tau_{i} \mathbf{e}_{z}^{B} &,
    \end{aligned}
    \end{equation}

\noindent
where $\mathbf{F}^{B}$ and $\mathbf{M}^{B}$ are the force and moment represented in the body frame, respectively. $x_{\mathrm{CoG}}$ and $y_{\mathrm{CoG}}$ are the $x$- and $y$-coordinates of the drone's center of mass, and $\mathbf{e}_{x}^{B}$, $\mathbf{e}_{y}^{B}$, and $\mathbf{e}_{z}^{B}$ are unit vectors of body $x$-, $y$-, and $z$-axes, respectively.
(\ref{eq4}) and (\ref{eq5}) are combined and expressed as follows:
    \begin{equation}
    \label{eq6}
        \mathbf{f}=\mathbf{A}^{-1} \mathbf{T},
    \end{equation}

\noindent    
where
    \noindent
    \begin{equation} \nonumber
        \mathbf{f} = \begin{bmatrix} f_1 & f_2 & f_3 & f_4 \end{bmatrix}^{\mathrm{T}},
    \end{equation}
    
    \begin{equation} \nonumber \scriptsize
        \mathbf{A}
        =
        \begin{bmatrix}
        1 & 1 & 1 & 1 
        \\
        -y_{1}+y_{\mathrm{CoG}} & -y_{2}+y_{\mathrm{CoG}} & -y_{3}+y_{\mathrm{CoG}} & -y_{4}+y_{\mathrm{CoG}} 
        \\
        x_{1}-x_{\mathrm{CoG}} & x_{2}-x_{\mathrm{CoG}} & x_{3}-x_{\mathrm{CoG}} & x_{4}-x_{\mathrm{CoG}} 
        \\
        \kappa / \mu & -\kappa / \mu & \kappa / \mu & -\kappa / \mu
        \end{bmatrix},
    \end{equation}

    \begin{equation} \nonumber
        \mathbf{T} = \begin{bmatrix} -F_z^{B} & M_x^{B} & M_y^{B} & M_z^{B} \end{bmatrix}^{\mathrm{T}}.
    \end{equation}

\section{Optimal Morphology} \label{sec3}

To determine the optimal morphology for energy-efficient and stable flight of the morphing quadrotor, the energy efficiency factor and controllability factor are defined, and the morphology maximizing these two factors is calculated. If only one of the two is considered, the optimal morphology is not uniquely determined. Because the morphing quadrotor is a drone that changes the angle of each arm, calculating the optimal morphology means calculating the optimal angle $\theta_{i}$ $( i=1,2,3,4)$ of each arm.

\subsection{Energy-Efficiency Factor}

An ``energy-efficient'' drone can be defined as a drone that consumes less energy when it generates the thrust for flight. If a drone is energy efficient, it can fly for a longer time, fly longer distances, and perform more tasks. We defined the energy-efficiency factor as the sum of the thrust of the rotor divided by the sum of the power consumed by each rotor when drone is hovering. Using the relationship (\ref{eq3}) between the thrust and power consumption of the motor, energy-efficiency factor $\eta$ is expressed as follows:

    \begin{equation}
        \label{eq7}
        \eta
        =
        \frac{\sum_{i=1}^{4} f_{i}}{\sum_{i=1}^{4} P_{i}}
        =
        \frac{\sum_{i=1}^{4} f_{i}}{\sum_{i=1}^{4} \gamma \sqrt{\frac{f_{i}^{3}}{2 \rho S}}}.
    \end{equation}

For the drone to hover, the sum of the thrust and gravity acting on the drone must be in equilibrium, and torque balance on the $x$-, $y$-, and $z$-axes must be established. Thus, the thrust $f_i$ $(i=1,2,3,4)$ of each rotor that must be generated for the drone to hover is calculated from (\ref{eq6}) by setting $\mathbf{T}=\begin{bmatrix}mg & 0 & 0 & 0\end{bmatrix}^{\mathrm{T}}$, where $m$ is the drone's mass and $g$ is the gravitational acceleration. $\mathbf{A}$ is the matrix determined by the positions of the rotors and the position of the drone's center of gravity, and the positions of the rotors are determined by morphology. Therefore, the energy-efficiency factor is determined by morphology if the drone's weight and center of gravity position are given.

\subsection{Controllability Factor}

The drone's control is accomplished by following the angular acceleration command $\boldsymbol{\alpha}_\mathrm{des}$ generated by the controller by changing the rotor thrust. Therefore, in order for the drone to be stably controlled, when the $\boldsymbol{\alpha}_\mathrm{des}$ changes, the drone must be able to quickly follow $\boldsymbol{\alpha}_\mathrm{des}$ through the thrust change of the rotor. In particular, it is important for stability and maneuverability of the drone to quickly follow angular acceleration command for roll and pitch, i.e., $\alpha_{x,\mathrm{des}}$ and $\alpha_{y,\mathrm{des}}$, through the thrust change of the rotor. If the required thrust of each rotor to follow the change of $\boldsymbol{\alpha}_\mathrm{des}$ changes significantly, it becomes difficult to follow $\boldsymbol{\alpha}_\mathrm{des}$ by changing the actual rotational speed of the rotor.

Therefore, we defined the controllability factor $C$ as the reciprocal of the max value among the vector magnitudes obtained by partial differentiation of the required thrust $f_{i}$ of each rotor with $\alpha_{x,\mathrm{des}}$ and $\alpha_{y,\mathrm{des}}$ as follows:

    \begin{equation}
        \label{eq8}
        C
        =
        \frac{1}{\underset{i}{\max} \left(\left\|\left[\frac{\partial f_{i}}{\partial \alpha_{x}, \mathrm{des}} \quad \frac{\partial f_{i}}{\partial \alpha_{y, \mathrm{des}}}\right]\right\|\right)}.
    \end{equation}

\noindent
The vector obtained by partial differentiation of $f_{i}$ with respect to $\alpha_{x,\mathrm{des}}$ and $\alpha_{y,\mathrm{des}}$ can be calculated as follows:

    \begin{equation}
    \label{eq9}
    \begin{split}
        & \begin{bmatrix}
        \frac{\partial f_{i}}{\partial \alpha_{x, \mathrm{des}}} &
        \frac{\partial f_{i}}{\partial \alpha_{y, \mathrm{des}}}
        \end{bmatrix}
        \\
        & =
        \begin{bmatrix}
        \frac{\partial f_{i}}{\partial M_{x}^{B}} \frac{\partial M_{x}^{B}}{\partial \alpha_{x, \mathrm{des}}}
        +
        \frac{\partial f_{i}}{\partial M_{y}^{B}} \frac{\partial M_{y}^{B}}{\partial \alpha_{x, \mathrm{des}}}
        +
        \frac{\partial f_{i}}{\partial M_{z}^{B}} \frac{\partial M_{z}^{B}}{\partial \alpha_{x, \mathrm{des}}}
        \\
        \frac{\partial f_{i}}{\partial M_{x}^{B}} \frac{\partial M_{x}^{B}}{\partial \alpha_{y, \mathrm{des}}}
        +
        \frac{\partial f_{i}}{\partial M_{y}^{B}} \frac{\partial M_{y}^{B}}{\partial \alpha_{y, \mathrm{des}}}
        +
        \frac{\partial f_{i}}{\partial M_{z}^{B}} \frac{\partial M_{z}^{B}}{\partial \alpha_{y, \mathrm{des}}}
        \end{bmatrix}^{\mathrm{T}}
        \\
        & =
        \begin{bmatrix}
        \frac{\partial f_{i}}{\partial M_{x}^{B}} & 
        \frac{\partial f_{i}}{\partial M_{y}^{B}} & 
        \frac{\partial f_{i}}{\partial M_{z}^{B}}
        \end{bmatrix}
        \begin{bmatrix}
        \frac{\partial M_{x}^{B}}{\partial \alpha_{x, \mathrm{des}}} &
        \frac{\partial M_{x}^{B}}{\partial \alpha_{y, \mathrm{des}}} \\
        \frac{\partial M_{y}^{B}}{\partial \alpha_{x, \mathrm{des}}} &
        \frac{\partial M_{y}^{B}}{\partial \alpha_{y, \mathrm{des}}} \\
        \frac{\partial M_{z}^{B}}{\partial \alpha_{x, \mathrm{des}}} &
        \frac{\partial M_{z}^{B}}{\partial \alpha_{y, \mathrm{des}}}
        \end{bmatrix}
        \\
        & = 
        \begin{bmatrix}
        \frac{\partial f_{i}}{\partial M_{x}^{B}} & 
        \frac{\partial f_{i}}{\partial M_{y}^{B}} & 
        \frac{\partial f_{i}}{\partial M_{z}^{B}}
        \end{bmatrix}
        \begin{bmatrix}
        J_{xx} & -J_{xy} \\
        J_{yx} & J_{yy} \\
        J_{zx} & -J_{zy}
        \end{bmatrix},
        \end{split}
        \end{equation}

\noindent
where inertia tensor $\mathbf{J}$ of the drone is epressed as follows:

    \begin{equation}
        \nonumber
        \mathbf{J} = \begin{bmatrix}
        J_{xx} & -J_{xy} & -J_{xz} \\
        J_{yx} & J_{yy} & -J_{yz} \\
        J_{zx} & -J_{zy} & J_{zz} 
        \end{bmatrix}.
    \end{equation}

\noindent
$\begin{bmatrix}\frac{\partial f_{i}}{\partial M_{x}^{B}} & \frac{\partial f_{i}}{\partial M_{y}^{B}} & \frac{\partial f_{i}}{\partial M_{z}^{B}}\end{bmatrix}$ can be obtained by partial differentiation of the result of equation (\ref{eq6}).

\subsection{Optimal Morphology}

\begin{figure}[t] 
    \centering
        \subfigure[]
        {
            \includegraphics[width=4cm]{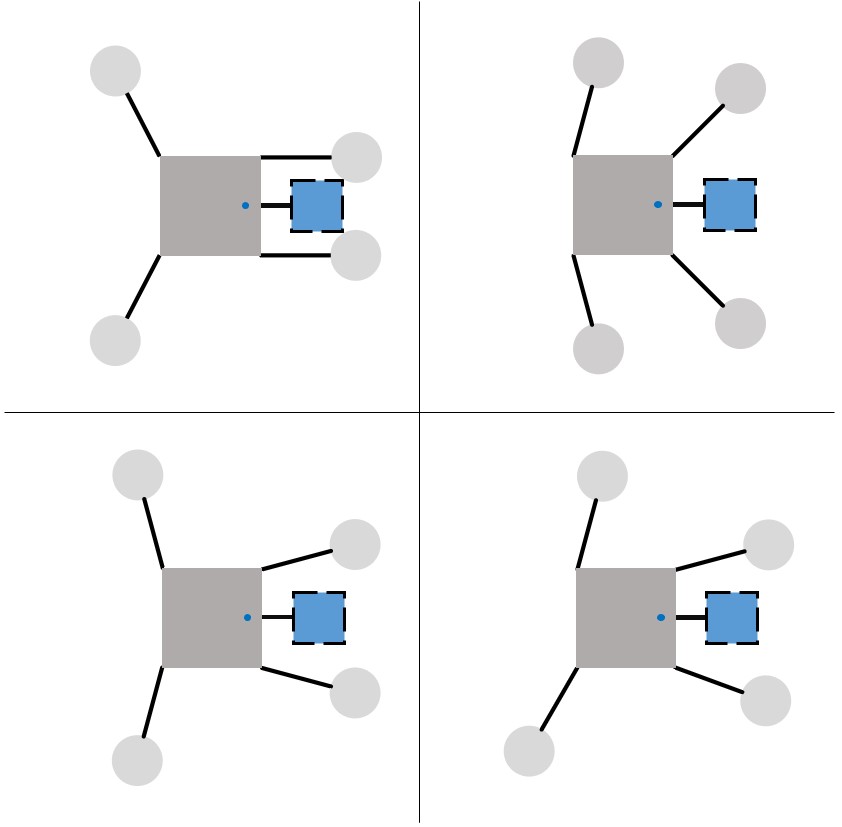}
            \label{fig_morphologyE}
        }
        \subfigure[]
        {
            \includegraphics[width=4cm]{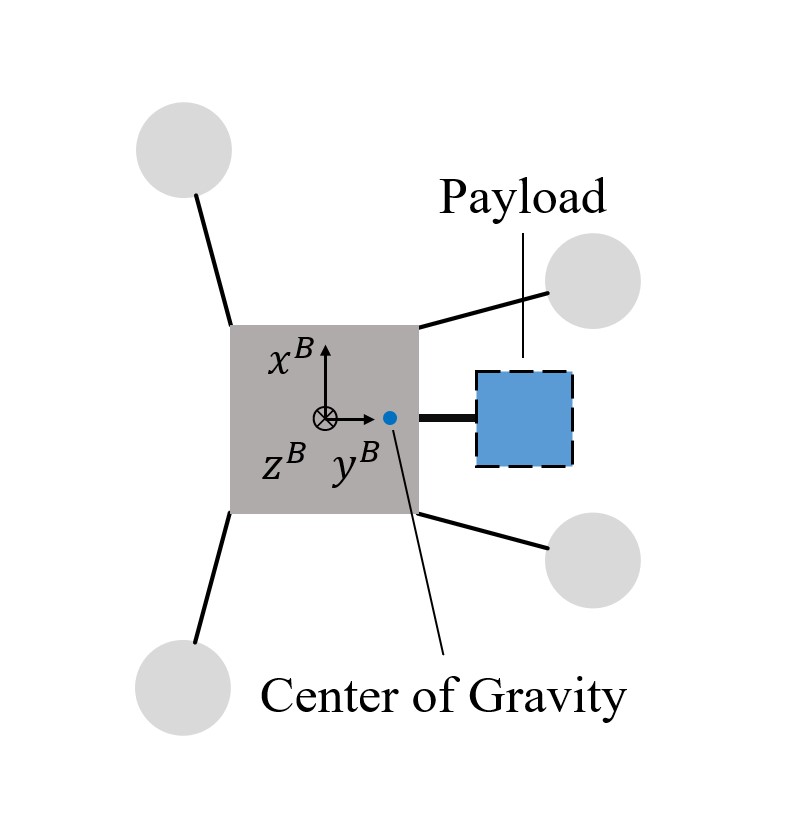}
            \label{fig_morphologyEC}
        }
      \caption{(a) Various morphology that maximizes $\eta$ exist. (b) The optimal morphology is uniquely determined by considering both $\eta$ and $C$.}
      \label{fig_morphologyE&EC}
\end{figure}

The energy-efficiency factor $\eta$ directly affects the flight time and flight distance of the drone, while the controllability factor $C$ only needs to be large enough to ensure a stable flight of the drone. Hence, we defined the optimal morphology as the morphology that maximizes $C$ among the morphologies that maximize $\eta$. By considering both $\eta$ and $C$, we can uniquely determine the optimal morphology. Because $C$ and $\eta$ are determined by morphology $(\theta_1, \theta_2, \theta_3, \theta_4)$, it is possible to represent $C$ and $\eta$ as functions of $(\theta_1, \theta_2, \theta_3, \theta_4)$. 

The gradient ascent algorithm was used to calculate the optimal morphology iteratively. By subtracting the vector component of $\nabla C$ parallel to $\nabla \eta$, $(\theta_1, \theta_2, \theta_3, \theta_4)$ will converge to a morphology that maximizes $C$ among morphologies that maximize $\eta$. The gradient ascent iteration between time steps $k$ and $k+1$ is obtained as follows: 

    \begin{equation}
    \label{eq10}
    \begin{split}
        (\theta_{1,k+1}, \theta_{2,k+1}, \theta_{3,k+1}&, \theta_{4,k+1})
        \\
        =
        (\theta_{1,k}, \theta_{2,k}&, \theta_{3,k}, \theta_{4,k})
        + \beta_{1} \nabla \eta_{k} \\
        & + \beta_{2}(\nabla C_{k} - \mathrm{proj}_{\nabla \eta_{k}} \nabla C_{k}),
    \end{split}
    \end{equation}
    
\noindent
where
    \begin{equation} \nonumber
    \begin{split}
        \nabla \eta_{k}
        & =
        \nabla \eta(\theta_{1,k}, \theta_{2,k}, \theta_{3,k}, \theta_{4,k}),
        \\
        \nabla C_{k}
        & =
        \nabla C(\theta_{1,k}, \theta_{2,k}, \theta_{3,k}, \theta_{4,k}),
    \end{split}
    \end{equation}
$\beta_1$ and $\beta_2$ are ascent rate constants for the gradient ascent, and $\mathrm{proj}_{\nabla \eta_{k}} \nabla C_{k}$ is the vector projection of $\nabla C_{k}$ on $\nabla \eta_{k}$.
    
\section{Control} \label{sec4}

\subsection{Parameter Estimation} 

When a payload is loaded on a drone, the drone's weight and center of gravity position change. The weight and center of gravity position are used for optimal morphology calculation and control allocation, so these parameters must be estimated and updated in real time. 

Assuming the drone is near the hovering state, $\mathbf{T} = \begin{bmatrix}-F_z^{B} & M_x^{B} & M_y^{B} & M_z^{B}\end{bmatrix}^{\mathrm{T}} \cong \begin{bmatrix}mg & 0 & 0 & 0\end{bmatrix}^{\mathrm{T}}$ through the thrust generated by each rotor, the weight $m$ of the drone and the position of the center of gravity $(x_\mathrm{CoG},y_\mathrm{CoG})$ are estimated as follows:

    \begin{equation}
        \label{eq11}
        mg = \sum_{i=1}^{4} f_{i},
    \end{equation}
    
    \begin{equation}
        \label{eq12}
        x_\mathrm{CoG} = \frac{\sum_{i=1}^{4} x_{i}f_{i}}{\sum_{i=1}^{4} f_{i}},\
        y_\mathrm{CoG} = \frac{\sum_{i=1}^{4} y_{i}f_{i}}{\sum_{i=1}^{4} f_{i}}.
    \end{equation}

The mass of the load is estimated as follows:
    \begin{equation}
        \label{eq11-1}
        m_\mathrm{load} = m - (m_\mathrm{body} + 4m_\mathrm{arm}),
    \end{equation}
\noindent
where $m_\mathrm{load}$, $m_\mathrm{body}$, and $m_\mathrm{arm}$ are masses of the load, main body and arm, respectively. 

When the payload is loaded, not only the weight and center of gravity of the drone change, but also the total inertia tensor $\mathbf{J}$. For the rate controller to generate an appropriate moment setpoint, $\mathbf{J}$ must also be estimated and updated in real time. $\mathbf{J}$ of the drone can be obtained by adding the inertia tensor of the main body, the inertia tensor of the four arms, and the inertia tensor of the payload as follows:

    \begin{equation}
    \label{eq13}
    \begin{split}
        \mathbf{J} = & \mathbf{J}_\mathrm{body} + \sum_{i=1}^{4} \mathbf{J}_{\mathrm{arm},i} + \mathbf{J}_\mathrm{load},
        \\
        \mathbf{J}_\mathrm{body} = & \mathbf{J}_\mathrm{body}^{cm} - m_\mathrm{body} [\mathbf{r}_\mathrm{body} - \mathbf{r}_\mathrm{CoG}]_{\times}^2,
        \\
        \mathbf{J}_{\mathrm{arm}, i}  = & \mathbf{R}_{z}(\theta_{i}) \mathbf{J}_\mathrm{arm}^{cm} \mathbf{R}_z(\theta_{i})^\mathrm{T}
        \\ 
        & - m_\mathrm{arm} [\mathbf{r}_{\mathrm{arm}, i} - \mathbf{r}_\mathrm{CoG}]_{\times}^2,
        \\
        \mathbf{J}_\mathrm{load} = & \mathbf{J}_\mathrm{load}^{cm} - m_\mathrm{load} [\mathbf{r}_\mathrm{load}-\mathbf{r}_\mathrm{CoG}]_{\times}^2,
    \end{split}
    \end{equation}
\noindent
where $\mathbf{J}_\mathrm{body}$, $\mathbf{J}_{\mathrm{arm},i}$, $\mathbf{J}_\mathrm{load}$ are inertia tensors of the main body, the $i$-th arm, and payload with reference to the drone's center of gravity, respectively. $\mathbf{J}_\mathrm{body}^{cm}$, $\mathbf{J}_\mathrm{arm}^{cm}$, $\mathbf{J}_\mathrm{load}^{cm}$ are inertia tensors of the main body, arm, and payload with reference to their own center of mass, respectively. $\mathbf{r}_\mathrm{body}$, $\mathbf{r}_{\mathrm{arm},i}$, $\mathbf{r}_\mathrm{load}$, $\mathbf{r}_\mathrm{CoG}$ are position vectors of the main body, the $i$-th arm, load, and drone's center of gravity. $[\cdot]_{\times}$ denotes skew-symmetric matrix. The $z$-coordinate of the center of gravity $z_\mathrm{CoG}$ is assumed to be the same as the $z$-coordinate of the drone's own center of gravity because it cannot be estimated from (\ref{eq12}). Because $\mathbf{J}_\mathrm{load}^{cm}$ is different depending on the shape of the payload, it is calculated assuming that the payload is a cube of uniform density, whose width is half of the width of the body.  $\mathbf{r}_\mathrm{load}$ can be calculated as follows: 

    \begin{equation}
    \label{equ14}
        \mathbf{r}_\mathrm{load} = \frac{m \mathbf{r}_\mathrm{CoG} -m_\mathrm{body}\mathbf{r}_\mathrm{body}-\sum_{i=1}^{4}m_\mathrm{arm}\mathbf{r}_{\mathrm{arm},i}}{m_\mathrm{load}}.
    \end{equation}

\subsection{Position and Attitude Control}

\begin{figure}[h] 
    \centering
        \includegraphics[width = 8cm]{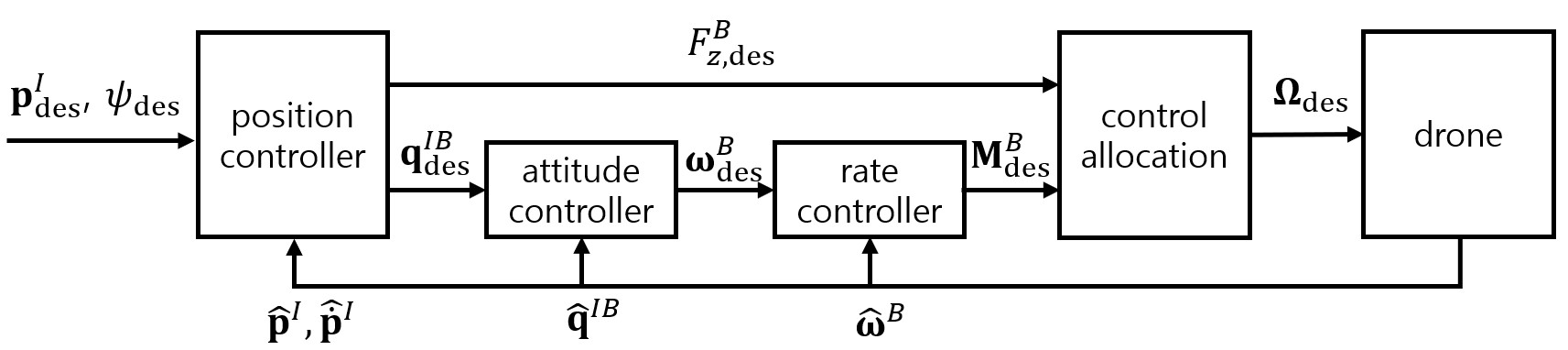}
        \caption{The control diagram of the morphing quadrotor}
        \label{fig3}
    \end{figure}

Position and attitude control is constructed based on the quaternion control of \cite{Dario}. In the position controller, the force $\mathbf{F}_{\text {des}}^{I}$ in the inertial frame that the drone needs to generate to follow the desired position is calculated as follows:

    \begin{equation}
    \label{eq15}
        \mathbf{F}_{\text {des}}^{I}
        = K_{\mathrm{p}, \mathrm{p}} \mathbf{p}_{\mathrm{err}} + K_{\mathrm{d}, \mathrm{p}} \dot{\mathbf{p}}_{\mathrm{err}} + K_{\mathrm{i}, \mathrm{p}} \int \mathbf{p}_{\mathrm{err}} d t + m \mathbf{g},
    \end{equation}
\noindent
where $\mathbf{p}_{\mathrm{err}}$ is the position error, which is the difference between the desired position $\mathbf{p}_{\mathrm{des}}^{I}$ and the estimated current position $\hat{\mathbf{p}}^{I}$ in the inertial frame. $K_{\mathrm{p}, \mathrm{p}}$, $K_{\mathrm{d}, \mathrm{p}}$, and $K_{\mathrm{i}, \mathrm{p}}$ are the proportional, derivative, and
integral gains of the position PID controller, respectively.

Using $\mathbf{F}_{\mathrm{des}}^{I}$ and desired yaw $\psi_{\mathrm{des}}$, it is possible to determine the desired orientation of the drone and its rotation matrix $\mathbf{R}_{\mathrm{des}}^{IB}$ so that the direction of the thrust vector of the drone matches the direction of $\mathbf{F}_{\mathrm{des}}^{I}$ as follows:

    \begin{equation}
    \label{eq16}
    \text{\scriptsize$
        \mathbf{R}_{\mathrm{des}}^{IB}
        =
        \mathbf{R}_{z} \left(\psi_{\mathrm{des}}\right)\left(\mathbf{I}_{3} + \frac{\left[\mathbf{e}_{z}^{I} \times \hat{\mathbf{F}}_{\mathrm{des}}^{I}\right]_{\times}^{2}}{\left(1+\mathbf{e}_{z}^{I} \cdot \hat{\mathbf{F}}_{\mathrm{des}}^{I}\right)}+\left[\mathbf{e}_{z}^{I} \times \hat{\mathbf{F}}_{\mathrm{des}}^{I}\right]_{\times}\right)
    $},
    \end{equation}
\noindent
where $\mathbf{e}_{z}^{I}$ is a unit vector of the $z$-axis of the inertial frame.

The quaternion $\mathbf{q}_{\mathrm{des}}^{IB}$ of $\mathbf{R}_{\mathrm{des}}^{IB}$ is calculated as follows:
    
    \begin{equation}
    \label{eq17}
    \begin{split}
        \mathbf{q}_{\mathrm{des}}^{IB}
        & =
        q_{r}+q_{i} \mathbf{i} + q_{j} \mathbf{j} + q_{k} \mathbf{k},
        \\
        q_{r} & =\frac{1}{2} \sqrt{1+R_{11}^{I B}+R_{22}^{I B}+R_{33}^{I B}},
        \\
        q_{i} & =\frac{1}{4 q_{r}}\left(R_{32}^{I B}-R_{23}^{I B}\right),
        \\
        q_{j} & =\frac{1}{4 q_{r}}\left(R_{13}^{I B}-R_{31}^{I B}\right),
        \\
        q_{k} & =\frac{1}{4 q_{r}}\left(R_{21}^{I B}-R_{12}^{I B}\right),
    \end{split}
    \end{equation}

\noindent
where $\mathbf{i}$, $\mathbf{j}$, and $\mathbf{k}$ are the fundamental quaternion units and $R_{ij}^{IB}$ is the $i$-th row, $j$-th column element of $\mathbf{R}_{\mathrm{des}}^{IB}$.

The amount of thrust $F_{z, \mathrm{des}}^{B}$ that the drone needs to generate in the $z$-axis direction of the body frame can be calculated as follows:

    \begin{equation}
    \label{eq18}
    F_{z, \mathrm{des}}^{B} = \left\|\mathbf{F}_{\mathrm{des}}^{I}\right\|.
    \end{equation}

The attitude control consists of an attitude controller that calculates the desired angular rate $\boldsymbol{\omega}_{\mathrm{des}}$ through the desired orientation $\mathbf{q}_{\mathrm{des}}^{IB}$ and a rate controller that calculates the desired moment $\mathbf{M}_{\mathrm{des}}^{B}$ through $\boldsymbol{\omega}_{\mathrm{des}}$.

The quaternion error $\mathbf{q}_{\mathrm{err}}$ is calculated as follows:

    \begin{equation}
    \label{eq19}
        \mathbf{q}_{\mathrm{err}}
        =
        \mathbf{q}_{\mathrm{des}}^{IB} \otimes \hat{\mathbf{q}}^{IB *} 
        = 
        \left(\begin{array}{l}
        q_{\mathrm{w}, \mathrm{err}} \\
        \mathbf{q}_{\mathrm{v}, \mathrm{err}}
        \end{array}\right),
    \end{equation}
\noindent
where $\hat{\mathbf{q}}^{\mathrm{IB}}$ is the estimation of quaternion between the inetial frame and the body frame of the drone, and $\otimes$ is the Hamilton product. $q_{\mathrm{w}, \mathrm{err}}$ and $\mathbf{q}_{\mathrm{v}, \mathrm{err}}$ are the scalar part and vector part of $\mathbf{q}_{\mathrm{err}}$, respectively. 

The attitude controller calculates the desired angular rate $\boldsymbol{\omega}_{\mathrm{des}}$ as follows:

    \begin{equation}
    \label{eq20}
        \boldsymbol{\omega}_{\mathrm{des}}
        =
        K_{\mathrm{a}} \operatorname{sign}\left(q_{\mathrm{w}, \mathrm{err}}\right) \mathbf{q}_{\mathrm{v}, \mathrm{err}},
    \end{equation}
\noindent
where $K_{\mathrm{a}}$ is the gain of the attitude controller and $\operatorname{sign}(\cdot)$ represents the sign function.

The rate controller calculates $\mathbf{M}_{\mathrm{des}}^{B}$ as follows:

    \begin{equation}
    \label{eq21}
    \text{\scriptsize$
        \mathbf{M}_{\mathrm{des}}^{B}
        =
        \mathbf{J}\left(K_{\mathrm{p}, \mathrm{r}} \boldsymbol{\omega}_{\mathrm{err}}+K_{\mathrm{d}, \mathrm{r}} \dot{\boldsymbol{\omega}}_{\mathrm{err}}+K_{\mathrm{i}, \mathrm{r}} \int \boldsymbol{\omega}_{\mathrm{err}} d t\right)+\hat{\boldsymbol{\omega}} \times \mathbf{J} \hat{\boldsymbol{\omega}}
    $},
    \end{equation}
\noindent
where $\boldsymbol{\omega}_{\mathrm{err}}$ is the difference between $\boldsymbol{\omega}_{\mathrm{des}}$ and the estimation of the drone's angular rate $\hat{\boldsymbol{\omega}}$; $K_{\mathrm{p}, \mathrm{r}}$, $K_{\mathrm{d}, \mathrm{r}}$, and $K_{\mathrm{i}, \mathrm{r}}$ are the proportional, derivative, and integral gains of the rate PID controller, respectively.

\subsection{Control Allocation}

Control allocation calculates the rotor speed creating desired force $F_{z, \mathrm{des}}^{B}$ and moments $M_{x, \mathrm{des}}^{B}$, $M_{y, \mathrm{des}}^{B}$, $M_{z, \mathrm{des}}^{B}$ generated from the position and attitude controllers. Unlike a general quadrotor, the position of the motor and the center of gravity change due to morphing and payload loading, so the control allocation matrix must be continuously updated in real time. Control allocation can be calculated as follows using (\ref{eq2}) and (\ref{eq6}):

    \begin{equation}
    \label{eq22}
    \begin{split}
        \boldsymbol{\Omega}
        & =
        \begin{bmatrix}
        \Omega_{1}^{2} & \Omega_{2}^{2} & \Omega_{3}^{2} & \Omega_{4}^{2}
        \end{bmatrix}^{\mathrm{T}}
        \\
        & =
        \mu^{-1} \mathbf{A}^{-1}
        \begin{bmatrix}
        F_{z, \mathrm{des}}^{B} & M_{x, \mathrm{des}}^{B} & M_{y, \mathrm{des}}^{B} & M_{z, \mathrm{des}}^{B}
        \end{bmatrix}^{\mathrm{T}}.
    \end{split}
    \end{equation}
    
\section{Experiment} \label{sec5}

\subsection{Hardware Description}

\begin{figure}[t]
    \centering
        \includegraphics[width = 8cm]{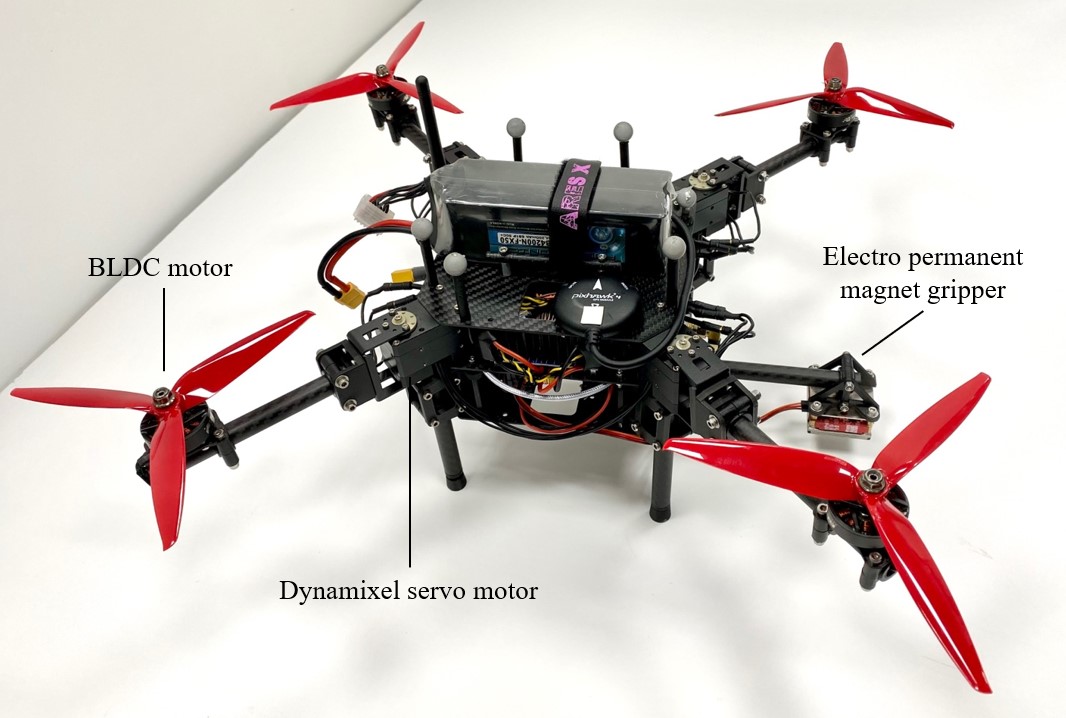}
        \caption{The components of our morphing quadrotor.}
        \label{fig_hardware}
\end{figure}

For the experiment, a morphing quadrotor weighing 2.0 kg with a gripper unit was used (Fig. \ref{fig_hardware}). The width of the main body is 120 mm and the length of each arms is 134 mm. The electrical and mechanical components of the morphing quadrotor are described in the following. 7 inch 3-blade propellers and T-Motor F90 1500KV brushless motors were used for the propulsion system, and a 6S 4200 mAh lithium polymer battery was used as a battery to supply power. Four Robotis Dynamixel XM430-W350-R servo motors were used to rotate four arms. Pixhawk4 mini \cite{Pixhawk4} was used as the flight controller, and Jetson Nano \cite{Jetson} was used as an onboard computer to calculate optimal morphology and to control the dynamixel servo motors. Opengrab Electro Permanent Magnet Gripper was used for payload gripping. The frame and other mechanical parts were made by carbon fiber machining and 3D printing.

We measured the power consumption of the rotor with motor test-bed and obtained the relationship between the thrust of the rotor $f_{i}$ [N] and the power consumption $P_{i}$ [W] as follows using (\ref{eq3}):

\begin{equation}
    \label{eq23}
        P_i 
        =
        8.9 \sqrt{f_{i}^{3}}.
    \end{equation}

\begin{figure}[t]
    \centering
        \subfigure[]
        {
            \includegraphics[width=4cm, height=4cm]{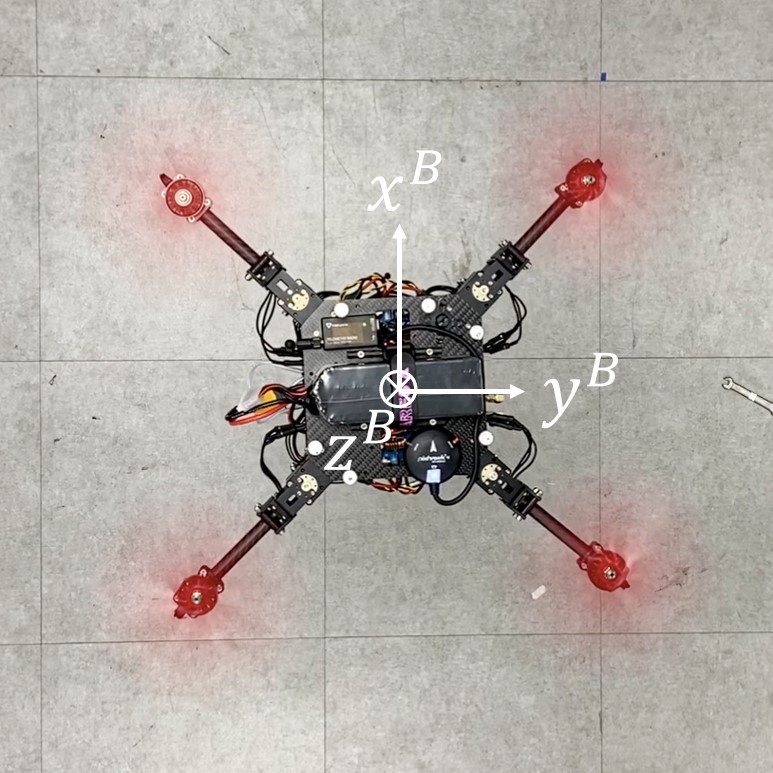}
            \label{fig_morphology_a}
        }
        \subfigure[]
        {
            \includegraphics[width=4cm, height=4cm]{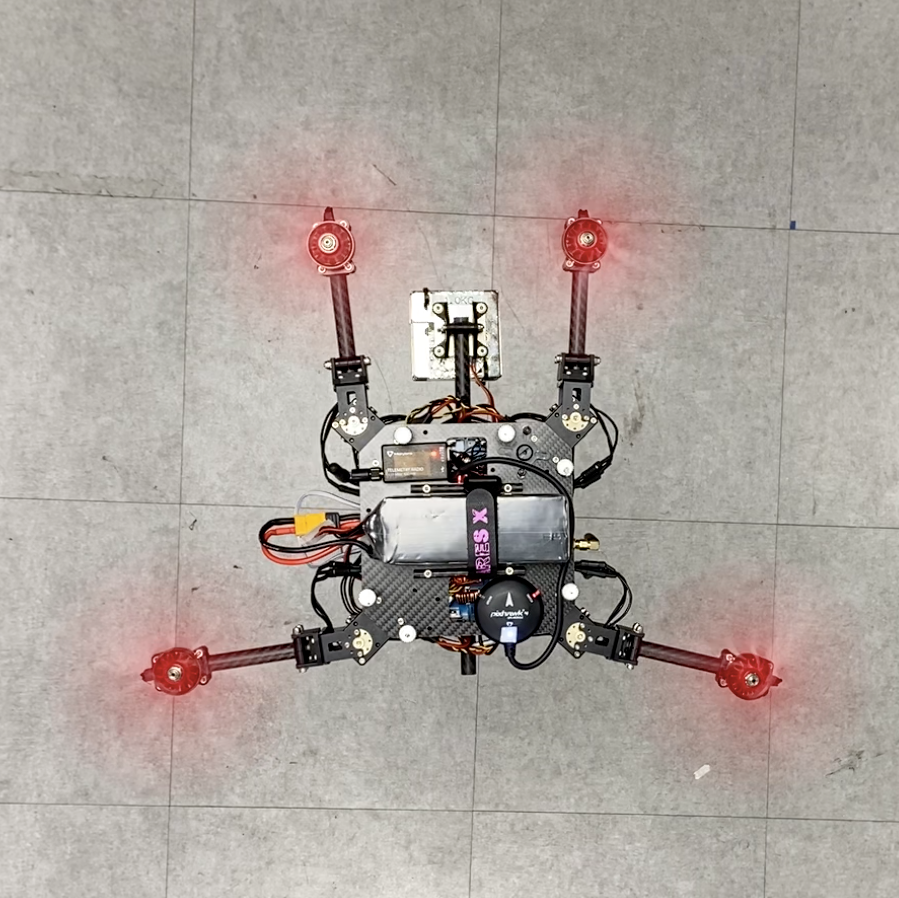}
            \label{fig_morphology_b}
        }
        \subfigure[]
        {
            \includegraphics[width=4cm, height=4cm]{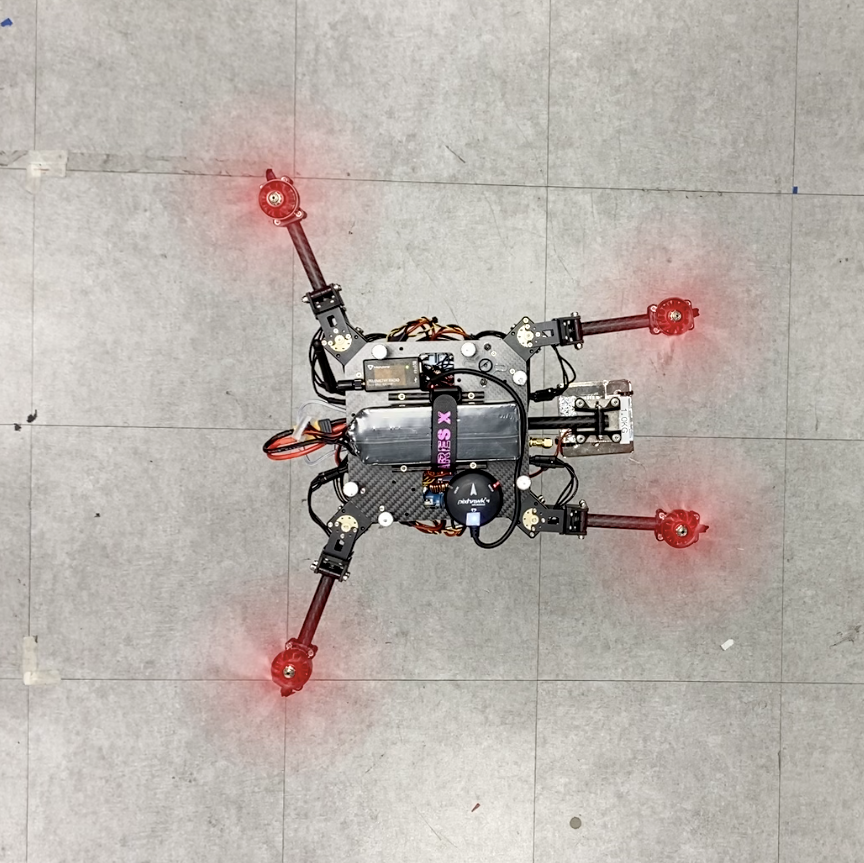}
            \label{fig_morphology_c}
        }
        \subfigure[]
        {
            \includegraphics[width=4cm, height=4cm]{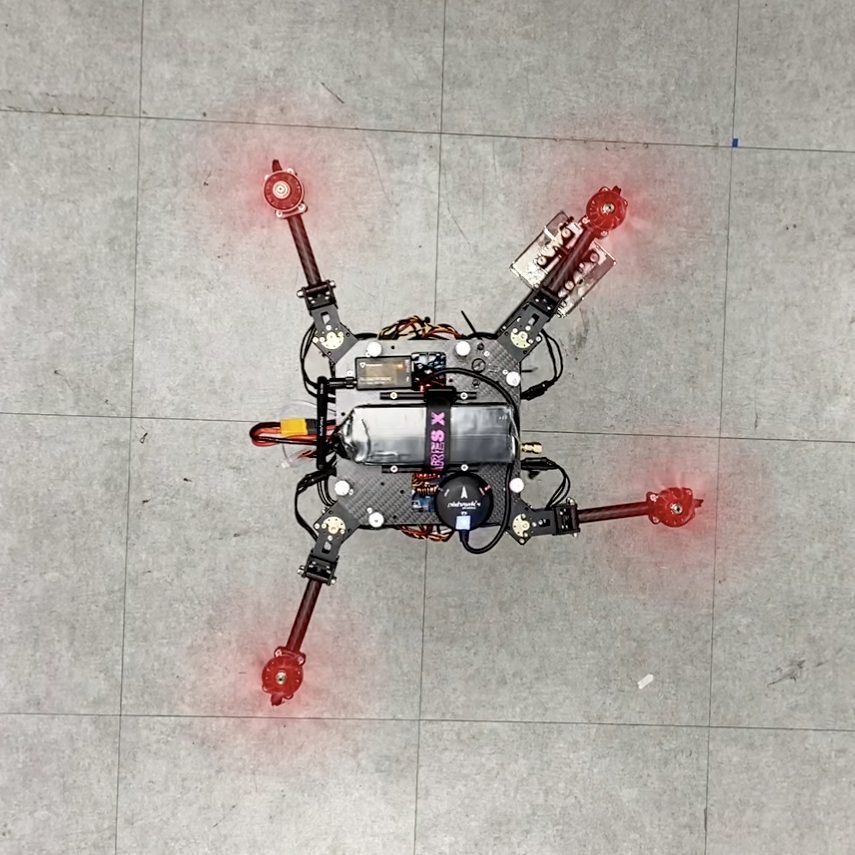}
            \label{fig_morphology_d}
        }
      \caption{The morphology of our morphing quadrotor while hovering with (a) 1 kg payload mounted at $x = 0$ cm, $y = 0$ cm, (b) at $x = + 15$ cm, $y = 0$ cm, (c) at $x = 0$ cm, $y = + 15$ cm, (d) at $x = 15/\sqrt{2}$ cm, $y = 15/\sqrt{2}$ cm.}
      \label{fig_morphology} 
\end{figure}

\begin{figure}[t]
    \centering
        \subfigure[]
        {
            \includegraphics[width=4cm]{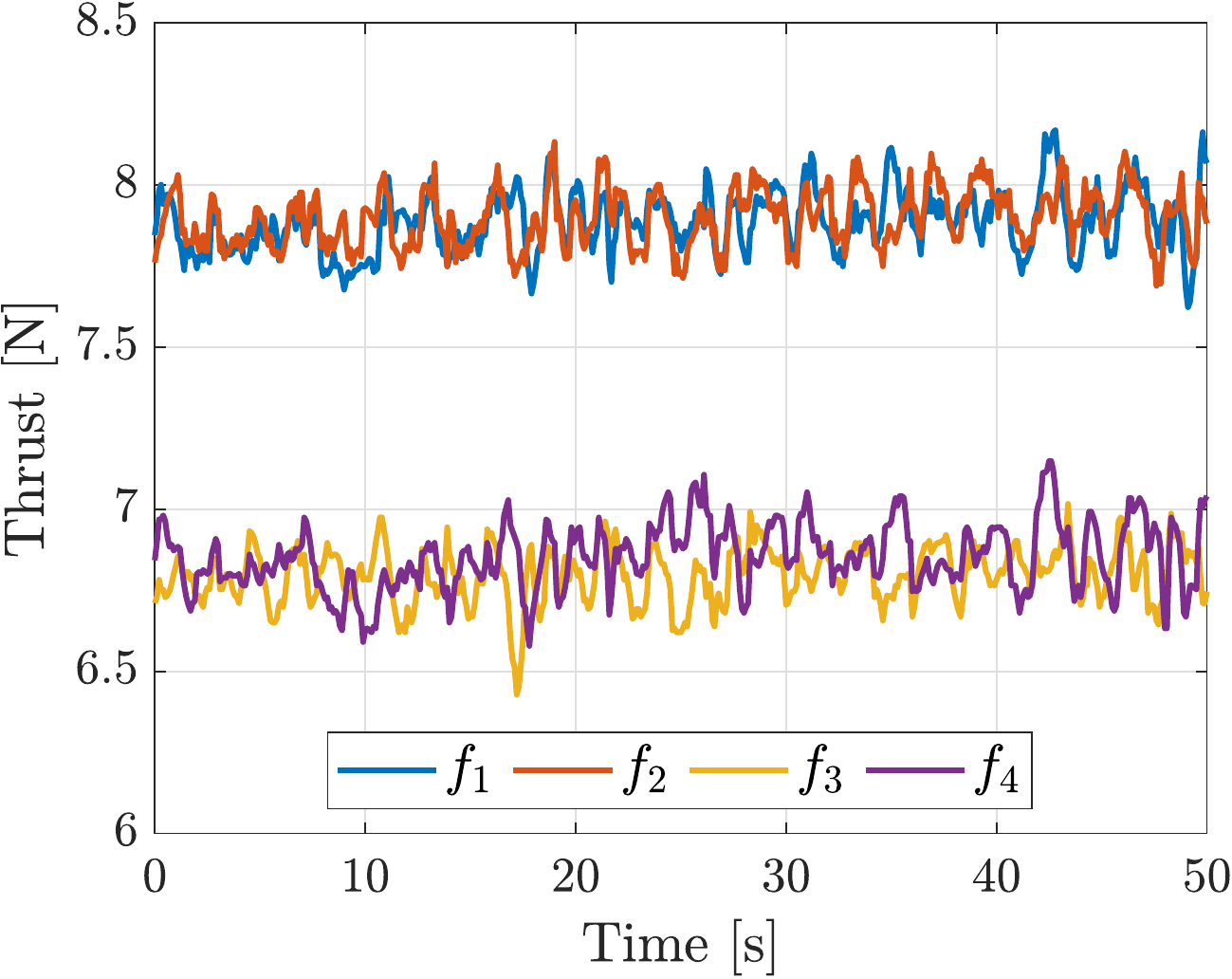}
            \label{fig_pwm_a}
        }
        \subfigure[]
        {
            \includegraphics[width=4cm]{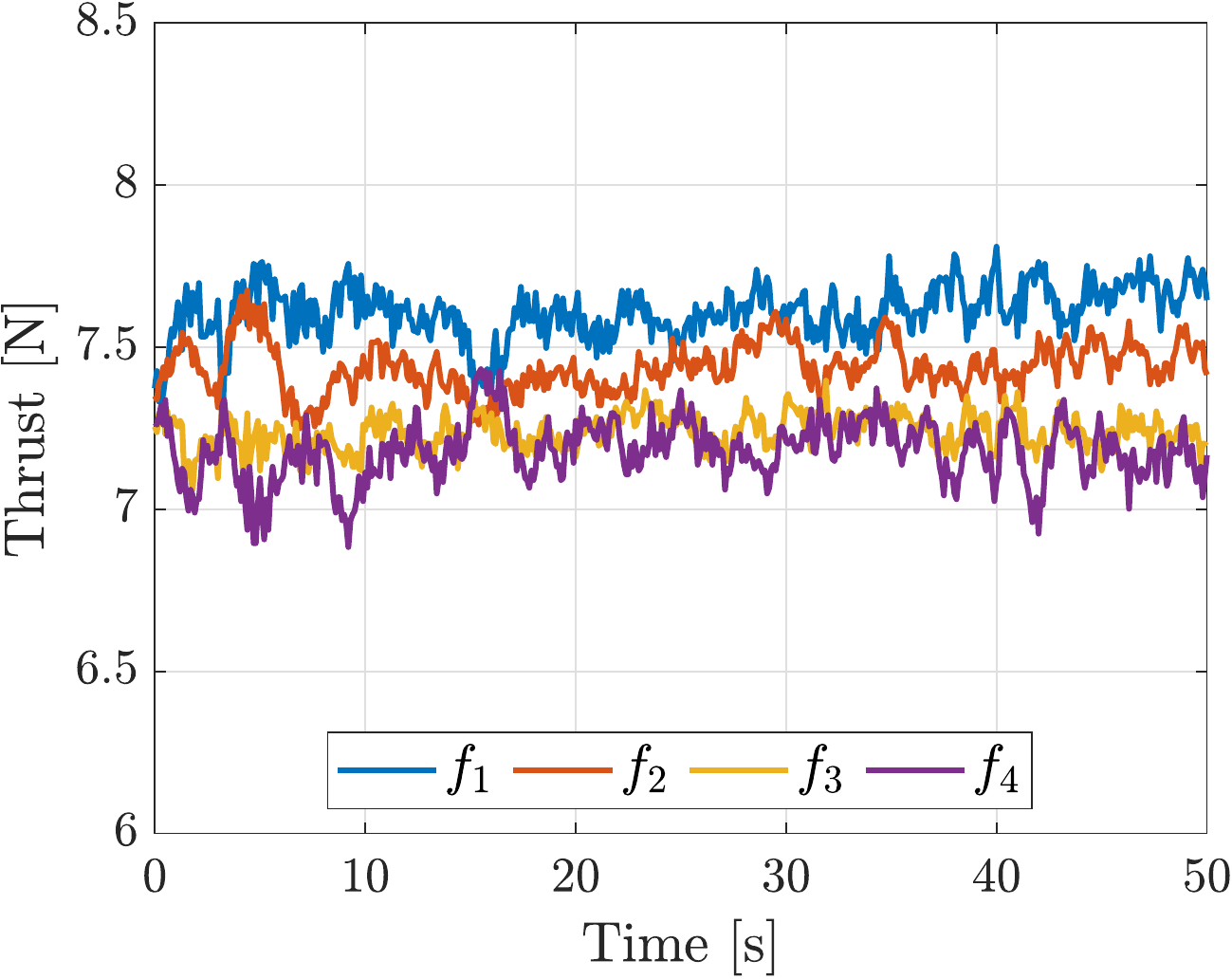}
            \label{fig_pwm_b}
        }
      \caption{Thrust of each rotor while hovering with 1 kg payload mounted at $x = 0$ cm, $y = +15$ cm.  (a) The conventional fixed-frame quadrotor. (b) Our morphing quadrotor.}
      \label{fig_pwm} 

\end{figure}

\subsection{Hovering}

For the flight test, a 1 kg payload, which is about 50\% of the weight of the drone, was mounted in various locations. The payload was installed at a distance of 0 cm, 5 cm, 10 cm, 15 cm, and 20 cm from the center of the drone in the $+x$, $+y$, and $xy$ diagonal directions. The test was conducted on both the morphing quadrotor and a conventional quadrotor whose frame is fixed to X configuration, and the results were compared.
\begin{figure}[!t]
    \centering
        \includegraphics[width = 8cm]{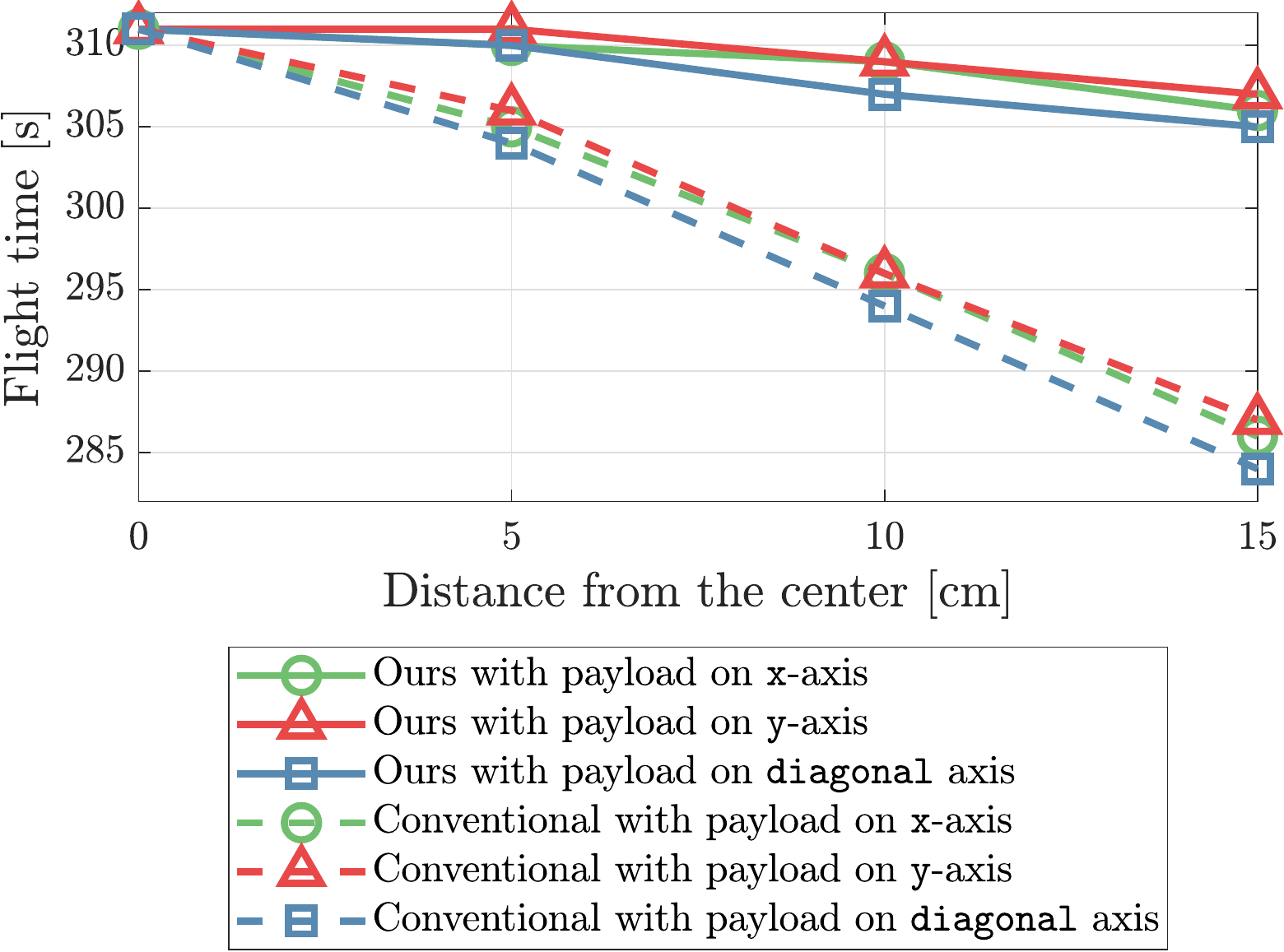}
        \caption{The flight time comparison between our morphing quadrotor and the conventional fixed-frame quadrotor}
        \label{fig_flighttime}
\end{figure}
\begin{figure}[!t]
    \centering
        \subfigure[]
        {
            \includegraphics[width=8cm]{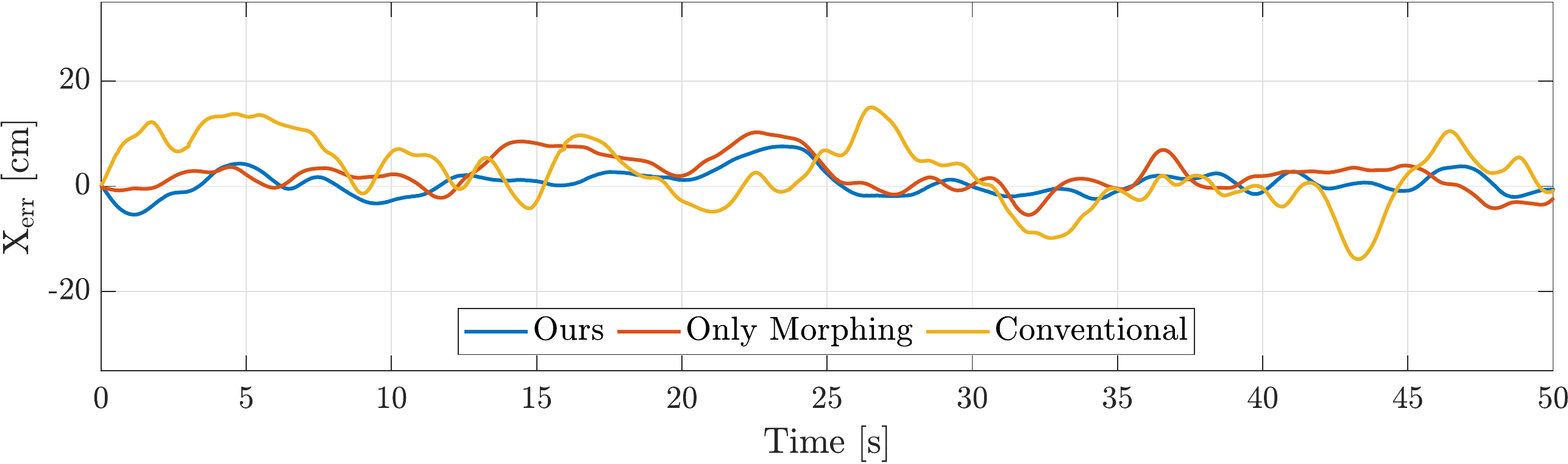}
            \label{fig_pos_a}
        }
        \subfigure[]
        {
            \includegraphics[width=8cm]{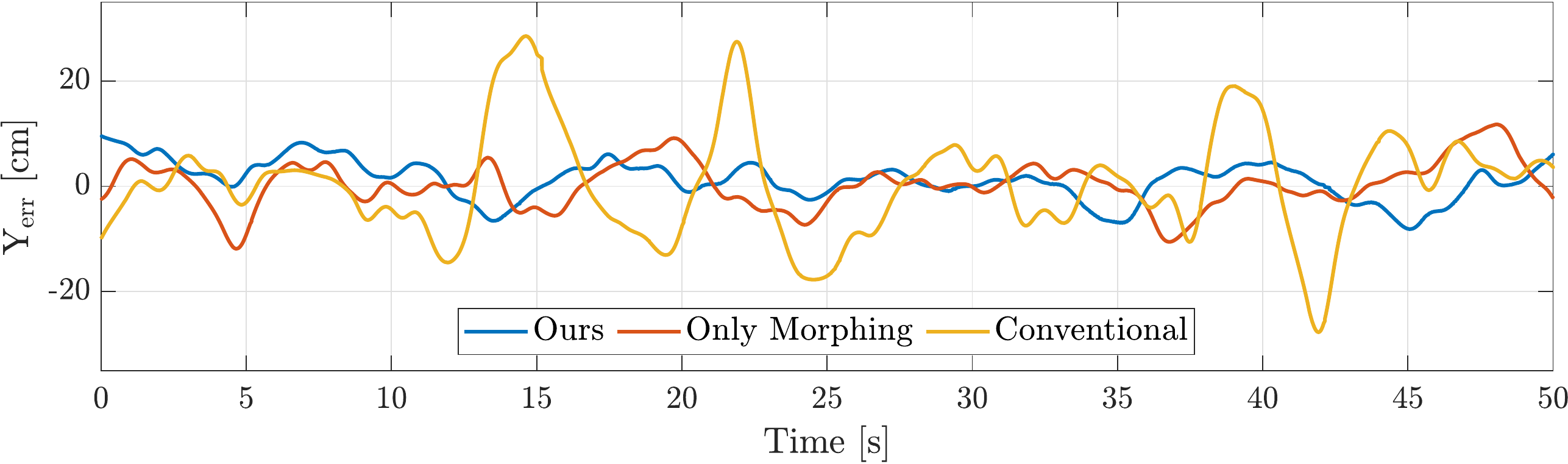}
            \label{fig_pos_b}
        }
        \subfigure[]
        {
            \includegraphics[width=8cm]{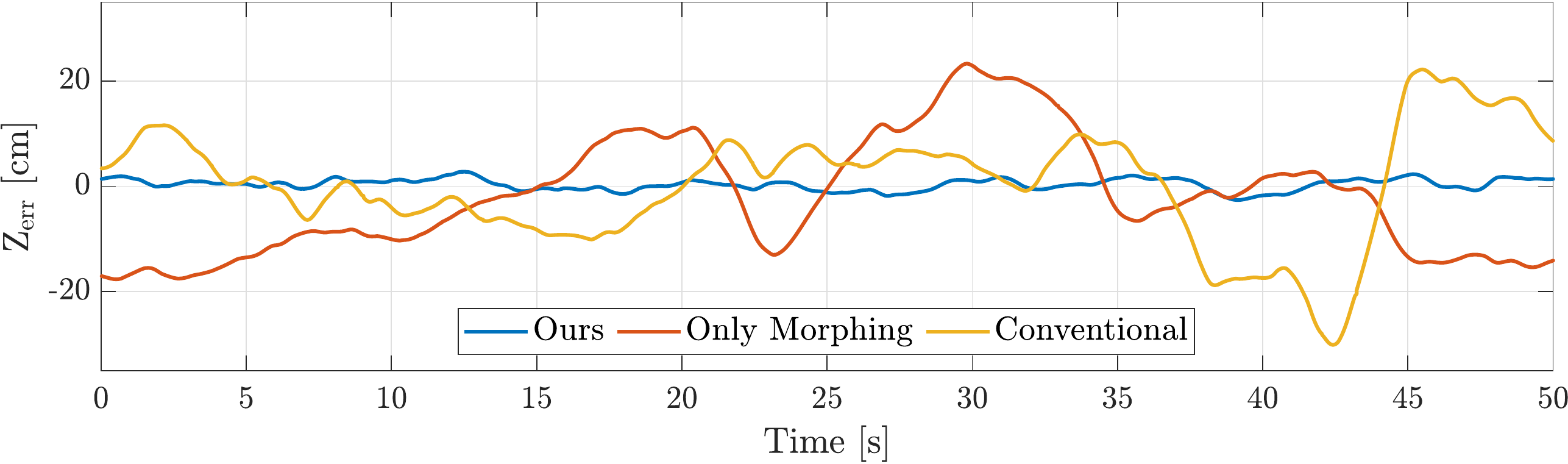}
            \label{fig_pos_c}
        }
      \caption{Position error while hovering with 1 kg payload mounted at $x = 0$ cm, $y = +15$ cm. Position error of our morphing quadrotor with adaptive controller (blue), the morphing quadrotor with conventional quadrotor controller (red), and the conventional fixed-frame quadrotor (yellow) are compared. (a) Position error in the $x$-direction. (b) Position error in the $y$-direction. (c) Position error in the $z$-direction.}
      \label{fig_pos} 

\end{figure}
In the case of the conventional quadrotor, the farther the payload was mounted from the center of the drone, the more unstable the flight was, and the difference in thrust between each rotor had to be increased to maintain hovering. It crashed when the payload was mounted at a distance of 20 cm from the center. On the other hand, as shown in Fig. \ref{fig_morphology}, our morphing quadrotor maintained stable flight, matching the center of thrust to the center of gravity by moving the rotor toward the payload.

Fig. \ref{fig_pwm} shows the thrust of each rotor during hovering. In the case of a conventional quadrotor, unbalanced rotor thrust must be generated to maintain hovering, but in the case of our morphing quadrotor, it is possible to hover by generating a uniform rotor thrust.  

According to (\ref{eq3}), the power consumption of the rotor increases rapidly as the thrust of the rotor increases. Thus, our morphing quadrotor can fly more efficiently than the conventional quadrotor, which has to generate large thrust from a specific rotor, resulting in high energy consumption. 

Fig. \ref{fig_flighttime} presents the result of comparing flight times when using 85\% of the battery. In the case of the conventional quadrotor, the flight time decreases as the payload moves away from the center, whereas the flight time of our morphing quadrotor remains almost constant.

Fig. \ref{fig_pos} shows the position error while hovering with 1 kg payload mounted at $x = 0$ cm, $y = +15$ cm. The conventional fixed-frame quadrotor showed unstable flight with large position error. The morphing quadrotor with the conventional quadrotor controller showed better performance compared to the conventional fixed-frame quadrotor, but position error was still large. Our morphing quadrotor with the adaptive controller maintained stable flight with small position error.

\subsection{Aerial Grasping and Dropping}

\begin{figure}[t] 
    \centering
    \includegraphics[width=8.5cm]{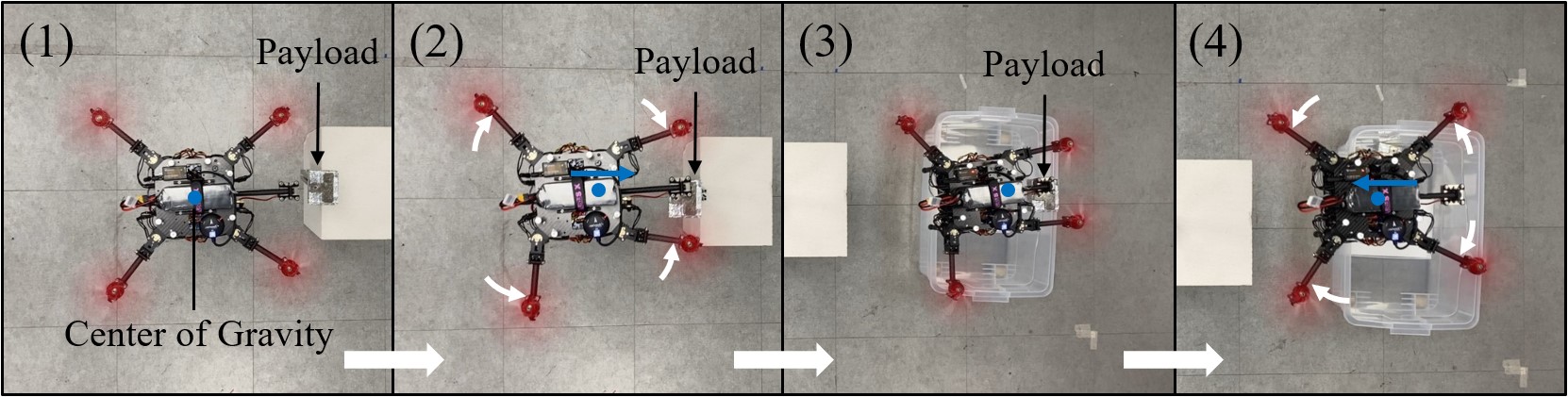}
    \caption{Aerial grasping and dropping experiment. (1) Approaching to the payload. (2) Grasping the payload. (3) Approaching to dropping point. (4) Dropping the payload. }
    \label{fig_grasp}
\end{figure}

\begin{figure}[t] 
    \centering
        \subfigure[]
        {
            \includegraphics[width=8cm]{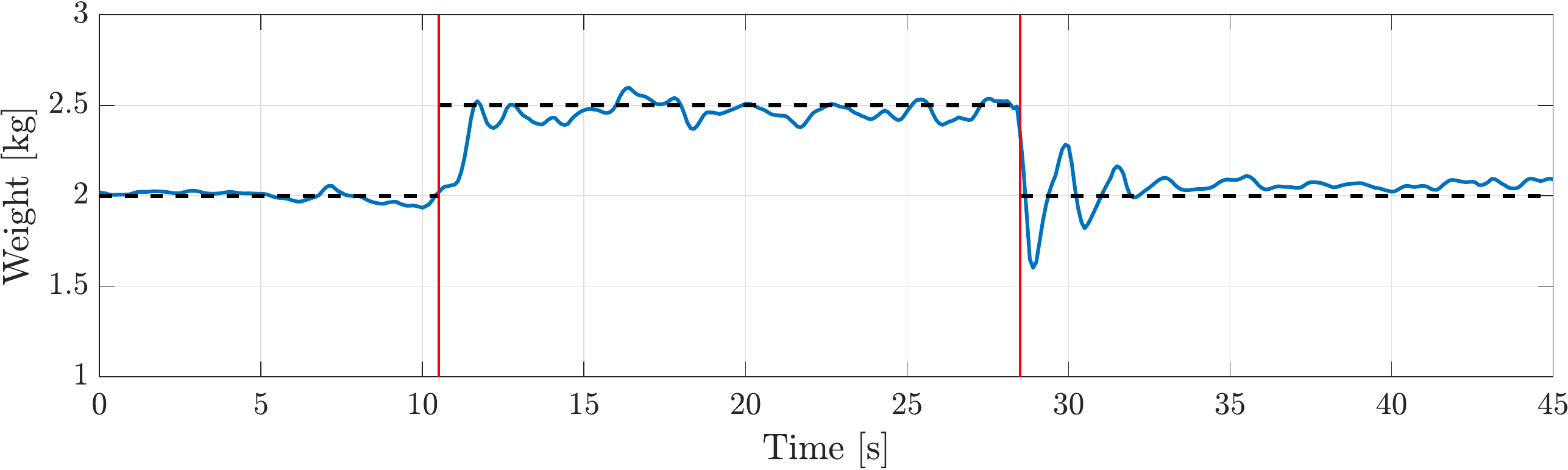}
            \label{fig_grasp_a}
        }
        \subfigure[]
        {
            \includegraphics[width=8cm]{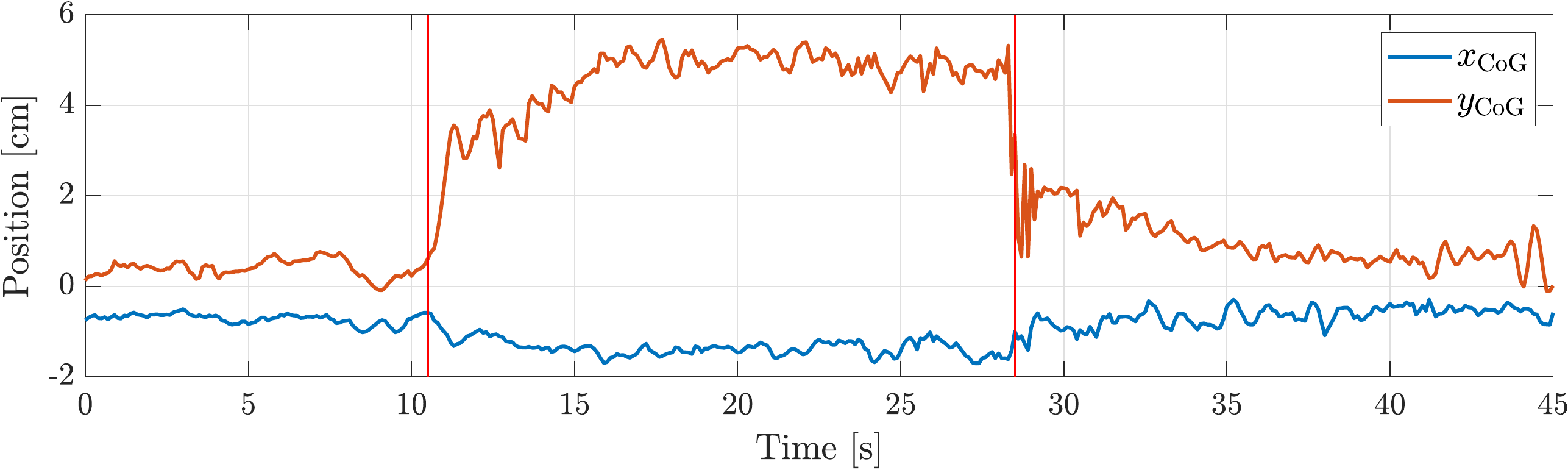}
            \label{fig_grasp_b}
        }
        \subfigure[]
        {
            \includegraphics[width=8cm]{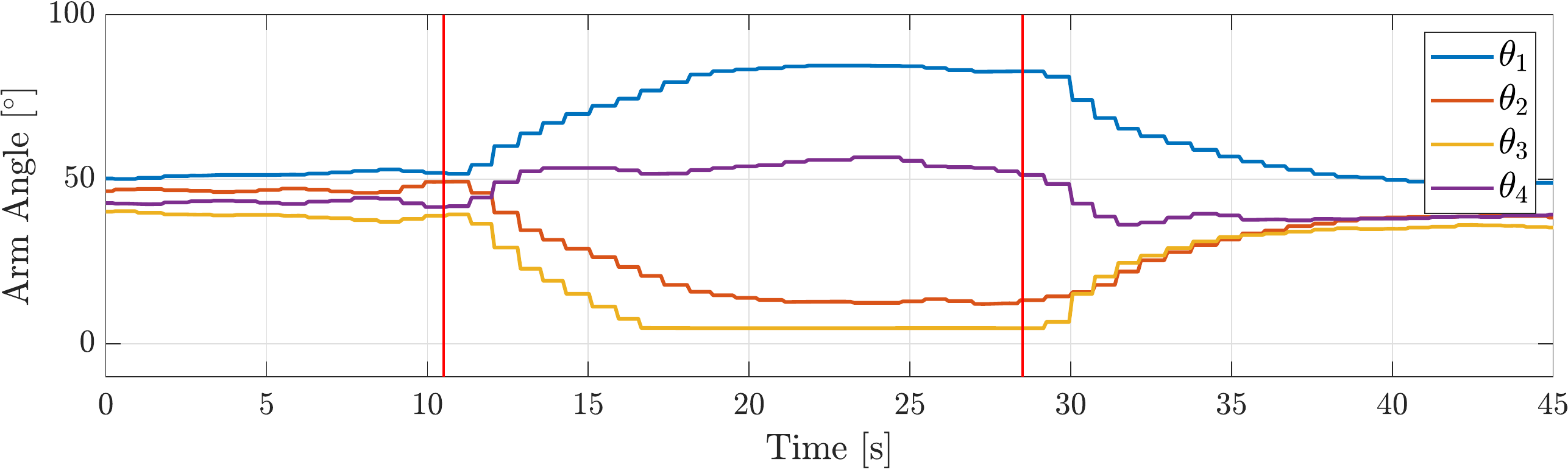}
            \label{fig_grasp_c}
        }
      \caption{Estimated parameters and arm angles in the aerial grasping and dropping experiment. The red vertical lines indicates the instants our morphing quadrotor grasps and drops the payload. (a) Total weight estimation. The actual weight is indicated with the dotted line. (b) Center of gravity estimation. (c) Arm angles. }
      \label{fig_graspgraph} 
\end{figure}

To demonstrate that the proposed morphing quadrotor effectively responds to changes in the center of gravity during flight, an aerial grasphing and dropping experiment was conducted. In the experiment, the task of grabbing an object weighing 0.5kg with a gripper unit mounted at $x = 0$ cm, $y = + 20$ cm and dropping it at a dropping point was given to our drone. As can be seen in Fig. \ref{fig_grasp} and Fig. \ref{fig_graspgraph}, while grasping the object, our morphing quadrotor estimated its changed weight and center of gravity position, and rotated its arms toward the object to maintain a stable flight. After dropping the object, it returned to the original X configuration and continued stable flight. 

\section{Conclusion} \label{sec6}

In this paper, we presented a novel method to improve the limitation on payload position and weight using a morphing quadrotor system. We defined the efficiency factor and controllability factor to evaluate the morphology and used these two factors to determine the optimal morphology for stable and efficient flight. When the payload is loaded onto the drone, the drone's weight, center of mass, and inertia tensor change. By estimating the drone's weight, center of gravity, and inertia tensor in real-time, the optimal morphology is updated in real-time, and the attitude controller and control allocation are updated according to the changed flight dynamics. By conducting flight tests in various situations and comparing with a conventional fixed-frame quadrotor, we demonstrated that the proposed morphing quadrotor and its control method improves the stability and efficiency for transporting the payload.





\end{document}